\documentclass[journal]{IEEEtran}

\usepackage{graphicx}
\usepackage{amsmath, amsfonts,epsfig, multirow, floatflt}
\usepackage{amssymb}
\usepackage{subfigure}
\usepackage{cite}
\usepackage{algorithm,algorithmic}
\usepackage{hyperref}
\usepackage{enumitem}
\usepackage{diagbox}
\usepackage{mathrsfs}
\usepackage{caption}
\usepackage{url}

\begin{document}
	
\title{Patent Analytics Based on Feature Vector Space Model: A Case of IoT}
	\author{\uppercase{Lei Lei}, 
	\uppercase{Jiaju Qi} and \uppercase{Kan Zheng}}
\maketitle
	\begin{abstract}
		The number of approved patents worldwide increases rapidly each year, which requires new patent analytics to efficiently mine the valuable information attached to these patents. Vector space model (VSM) represents documents as high-dimensional vectors, where each dimension corresponds to a unique term. While originally proposed for information retrieval systems, VSM has also seen wide applications in patent analytics, and used as a fundamental tool to map patent documents to structured data. However, VSM method suffers from several limitations when applied to patent analysis tasks, such as loss of sentence-level semantics and curse-of-dimensionality problems. In order to address the above limitations, we propose a patent analytics based on feature vector space model (FVSM), where the FVSM is constructed by mapping patent documents to feature vectors extracted by convolutional neural networks (CNN). The applications of FVSM for three typical patent analysis tasks, i.e.,  patents similarity comparison, patent clustering, and patent map generation are discussed. A case study using patents related to Internet of Things (IoT) technology is illustrated to demonstrate the performance and effectiveness of FVSM. The proposed FVSM can be adopted by other patent analysis studies to replace VSM, based on which various big data learning tasks can be performed.  
	\end{abstract}
	
	
	

\section{Introduction}

Patent analytics is a family of techniques and tools for analyzing the technological information presented within and attached to patents \cite{abbas2014literature}. A patent document contains various data such as title, abstract, application and filed dates, inventors' names, claims, figures, International Patent Classification (IPC) codes, and citations. Hence, patent data comprises diverse and plentiful results of various technologies. For example, we could explain the evolution of a technology by analyzing the number of relevant patents filed by year. As a technology is usually dependent on a group sub-technologies, using text mining techniques for patent keywords, we may understand the technological linkages between sub-technologies. Moreover, we could identify key technologies through patent citation analysis. Therefore, analysis, visualization and interpretation of data included in patent documents are very useful in technology innovation and forecast. However, the global patent data set is huge by any measure, with millions of new patent documents and updates made public every week. Weeding through this data to obtain useful information is a crucial but daunting task for IP professionals. Patent data is well suited for big data tools and techniques because of the volume, variety and velocity of changes. Therefore, how to apply big data analytics and learning to the patent industry for substantially better analyzing and visualizing patent data has been a hot research topic in recent years \cite{choi2018innovation,ki2017generating,zheng2015smdp,joung2017monitoring,nam2017monitoring,chen2017topic,liu2018patent,lei2016optimal,kim2016generating,trappey2016review,Venugopalan2015,song2017anticipation,wu2010method}. \par

The generic patent analysis workflow usually includes three steps \cite{abbas2014literature}. First, the patent documents related to the target technology are retrieved from patent databases. Next, the patent documents, which include a combination of structured and unstructured data, are transformed to structured data by employing text mining techniques. For example, the patent keywords related to the target technology can be extracted, based on which the patent-keyword matrix (structured data) is built. Finally, based on the structured data, the big data learning approaches including classification, regression, and clustering, etc., are used for various purposes, such as patent novelty detection and identifying patent quality, trend analysis and technology forecasting, managing R\&D planning, etc.. The visual output of the patent data can be represented in the form of graphs, networks and patent maps \cite{wu2016patent,cheng2014patent}.\par

In the patent analysis process, how to generate and extract useful information from a patent to accurately characterize its key features in a concise format is one of the key issues. The vector space models (VSMs), which have been widely used in information retrieval and text clustering, have been adopted to represent patents in many existing literature \cite{uchida2004patent,younge2016patent,choi2018innovation,chen2011ipc}. VSMs represent documents as vectors with multiple terms. There are different methods to construct VSMs, and one popular method used in recent studies is to generate a weighted vector for each patent based on the term-frequency of each term (i.e., keyword) for the patent, scaled by the inverse document-frequency of each term \cite{younge2016patent,Kim2015}. In other words, a patent is represented by a vector of the term-frequency-inverse-document-frequency (TF-IDF) weights of its keywords. In this way, all the patents can be projected in the vector space, and metrics such as the Euclidean distances between different patents can be used to quantify their similarities. \par

While the VSMs can disambiguate documents, they also have some limitations. As we know, there are two levels of semantics for a paragraph of sentences; i.e., word-level semantics and sentence-level semantics. The latter is more complex, comprehensive and high-level than the former. Specifically, VSMs are constructed based on terms or keywords, which are bound to result in losing more or less sentence-level semantics. Moreover, a large number of terms are usually needed to accurately reflecting the key patent information, which may result in the curse-of-dimensionality problem leading to a catastrophe for further mathematical analysis based on VSMs. Finally, while it is computationally easy to derive the TF-IDF weights, VSMs may suffer from information loss that limits the patent analysis performance, such as accuracy in identifying patents similarities.\par

In order to overcome the above limitations, in this paper, we propose a new model referred to as the patent feature vector space model (FVSM). The FVSM is a space of patent feature vectors, where the feature vector of a patent corresponds to the one obtained from the pooling layer of a convolutional neural network (CNN). Different from VSMs which have a term-based dimensional space, the dimension of FVSM is abstract and based on the neurons of CNN. This is inspired by the successful applications of CNN in extracting the feature vectors of images, and also its break-through results in some Natural Language Processing (NLP) tasks, e.g., sentence classification \cite{kim2014convolutional,young2018recent,zhang2017sensitivity}. Compared with the traditional VSMs, the proposed FVSM has the following three advantages. Firstly, the feature vector of a patent in FVSM is obtained by feeding sentences instead of only keywords as input data to the CNN, so that the sentence-level semantics can be captured. Second, by adjusting the number of neurons in the pooling layer of CNN, the space cardinality of FVSM can be flexibly controlled to avoid the curse-of-dimensionality problem. Finally, although the computational complexity of obtaining feature vectors of FVSM is larger than that of VSM, FVSM generally enjoys higher accuracy in characterizing the patent features thanks to the neural networks' internal ability in learning any non-linear functions. In this paper, we will demonstrate the superior performance of FVSM by applying it to several patent analysis tasks on the Internet-of-Things (IoT) technology, such as comparing patent-to-patent similarities and generating patent maps.\par 

The main contributions of this paper lie in the following two aspects. First, we propose a novel method of extracting feature vectors from patent documents using CNN. Although CNN has been successfully adopted for NLP tasks as well as extracting feature vectors of images, it has seldom been used to extract feature vectors of documents. Our work demonstrates the capability of CNN in this task. Second, the FVSM construction addresses a fundamental problem in patent analytics - efficiently mapping patent documents to structured data. The proposed FVSM can be adopted by other researchers, based on which various big data learning methods can be applied.    \par

The rest of the paper is organized as follows. The related works are summarized in Section II. In Section III, we provide a detailed description of the construction and application of the proposed FVSM. In Section IV, we perform patent analysis for IoT technology based on FVSM as a case study to demonstrate its performance. Finally, conclusion is given in Section V.

\section{Related Works}
\subsection{Vector Space Model}
In VSM, documents are expressed as vectors and the corpus of document is mapped into a high-dimensional space. VSM was originally developed for the SMART retrieval system \cite{Salton1975}, and has become one of the most robust methods in the field of Information Retrieval. The construction and application of VSM for patent analysis have been studied. In \cite{younge2016patent}, the entire USPTO patent space is mapped into a single vector space model. Specifically, a weighted vector for each patent document is generated based on the TF-IDF weights of each term for a patent (i.e., TF-IDF-based VSM). Based on this VSM, patent-to-patent similarity is calculated by the cosine of the angular separation between every two patents in the population. In \cite{Kim2015}, TF-IDF-based VSM is used for technology forecasting. Specifically, patent documents concerning similar technologies are clustered by K-means clustering method in VSM. Next, in order to assign a definition for each technology cluster, Latent Dirichlet Allocation (LDA) is used to extract latent topics from patent documents. LDA is a generative probabilistic model of a collection of documents made up of terms. It estimates two tables as final outputs. The first table describes the probability of selecting a particular term when sampling a particular topic. The second one describes the probability of selecting a particular topic when sampling a particular document. Finally, vacant and saturated technology clusters are identified according to the number of patents in the clusters. TD-IDF-based VSM is also used to generate the topic distribution table for each document by LDA in \cite{choi2018innovation}. Then, innovation topics and their relationship are identified by constructing innovation topic networks. One limitation of the above TF-IDF-based VSM is that co-occurrences of keywords are not considered. In order to address this problem, a keyword vector space based on word co-occurrences in close proximities in documents is obtained in \cite{uchida2004patent}. Then, the vector for a patent document is represented as the center of gravity with keyword vectors comprised from it. Finally, a patent map is generated from this VSM by clustering patent documents according to degree of similarity of their document vectors. However, all the above VSMs suffer from loss of sentence-level semantics and curse-of-dimensionality problems as discussed in Section I.\par

\subsection{Convolutional Neural Network for NLP}
CNN is one of the most influential innovations in the field of image processing. Since the pioneer work by Collobert \cite{Collobert2008}, CNN has also in the past years shown break-through results in some NLP tasks, e.g., sentence classification \cite{young2018recent}. A typical CNN structure consists of three types of layers from input to output - convolution, pooling and full-connection. For NLP tasks, word embeddings are often used as the first data processing layer in CNN as in \cite{kim2014convolutional,zhang2017sensitivity}. Typically, word embeddings are pre-trained by optimizing an auxiliary objective such as predicting a word based on its context. Therefore, the learned word vectors can capture context similarity and semantic information. The pre-trained word embedding can remain static or can be trained and fine-tuned with the other parameters of CNN.   \par

To the best of our knowledge, there are currently only few work on applying CNN for patent analysis. In \cite{venugopalan2015topic}, a neural network with one hidden layer is used for patent classification. In this paper, we will use CNN in a patent classification task. However, the purpose is to obtain the feature vector instead of the category for the patent.\par

\section{Proposed Method}
\subsection{FVSM Construction}

\subsubsection{CNN input}
First, patent data is collected based on a specified rule (e.g., having the same patent class code), and all the selected patents make up one data set. Then, we need to pre-process the patent documents and convert the unstructured data (e.g., title, abstract) into structured data by text mining techniques. The main task in the pre-processing stage is to convert a patent document into an eligible input for the CNN. Firstly, the stop words, punctuation and the short words in the title and abstract which cannot constitute compound words are removed, and the rest of words are lemmatized and stemmed. Secondly, the unique terms from all the patent documents in the data set are identified and included in a term dictionary. Consider there are $H$ unique terms in total, then each term is given an index ranging from $1$ to $H$. Finally, a patent document in the data set with $N$ terms is converted into an $N$-dimensional vector $\mathbf d=[t_{1},t_{2},\cdots,t_{N}]$, where $t_{n}$, $n\in\{1,2,\cdots,N\}$ is the index in the term dictionary of the $n$-th term in vector $\mathbf d$, i.e., $t_{n}\in\{1,2,\cdots,H\}$. Notice that the terms in vector $\mathbf d$ are arranged according to their sequence of appearance in the document, which means that the $n$-th term in $\mathbf d$ have $(n-1)$ unique terms appearing before it in the patent document. In this way, the sentence-level semantics can be preserved for later analytical stages by CNN.\par

%
%

\subsubsection{CNN architecture}
\begin{figure*}
	\centering
	\includegraphics[width=0.9\textwidth]{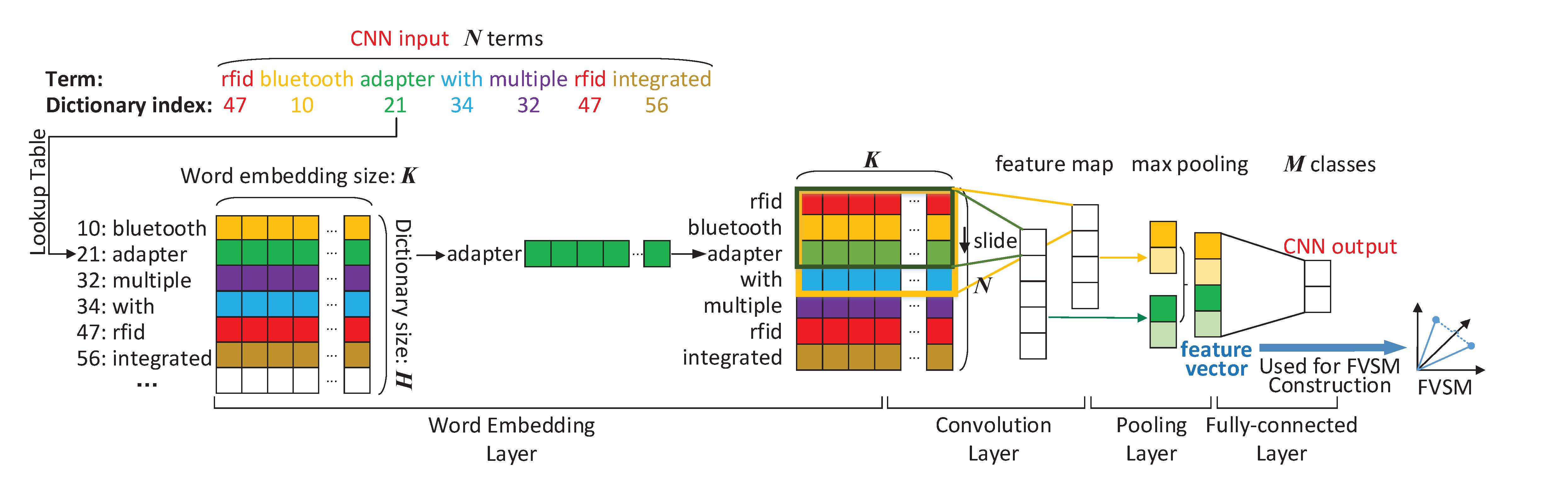}
	\caption{CNN architecture to extract the feature vector of a patent document.}
	\label{cnn}
\end{figure*}

In our proposed CNN as illustrated in Fig. \ref{cnn}, the first layer embeds terms of patent text into low-dimension vectors. The next two layers extract the advanced features from the vector. Then, the full-connection layer adds dropout regularization and classifies the result with a probability using a soft-max layer. The input to CNN is the $N$-dimensional vector $\mathbf d$ representing the text of a particular patent as discussed above. The output of CNN is the classification result of the patent. However, different from the typical NLP tasks such as sentence classification, our purpose in training CNN is to derive the feature vectors from the pooling layer for all the patents in the data set and use them to construct the FVSM. \par 


\paragraph{Word embedding layer}
The purpose of the word embedding layer is to convert the $N$-dimensional input vector $\mathbf d$ for a patent with $N$ terms into an $(N\times K)$-dimensional matrix $\mathbf{X}$, where the $n$-th row vector ($n\in\{1,\cdots,N\}$) in $\mathbf{X}$ corresponds to the $K$-dimensional word embedding of the $n$-th term in $\mathbf d$. In other words, each term in $\mathbf d$ is mapped to a $K$-dimensional vector by the word embedding layer.\par  

To realize the above mapping function, an $(H\times K)$-dimensional word embedding matrix $\mathbf{W}_{e}\in\mathbb{R}^{H\times K}$ is used as the weight matrix between the input layer and word embedding layer as shown in Fig. \ref{cnn}. The $h$-th row vector ($h\in\{1,\cdots,H\}$) represents the $h$-th term in the term dictionary. Therefore, for the $n$-th $(n\in\{1,\cdots,N\})$ term in the input vector $\mathbf d$, we retrieve its vector representation as the $t_{n}$-th row vector in the word embedding matrix $\mathbf{W}_{e}$. As a result, based on the $N$-dimensional input vector $\mathbf d$, we can construct the $(N\times K)$-dimensional matrix $\mathbf{X}$ as the output of the word embedding layer, which is used as input to the following convolutional layer for further processing. \par

Generally speaking, there are two approaches to obtain the word embedding matrix $\mathbf{W}_{e}$. The first approach is by the static method, where the word embeddings are directly obtained from an unsupervised neural language model, e.g., Word2vec, where the word embedding vectors were trained on $100$ billion words of Google News and are publicly available \cite{goldberg2014word2vec}. For the static method, as the weights between the input layer and word embedding layer, i.e., the word embedding matrix $\mathbf{W}_{e}$, are static and not trained with the other parameters of CNN, we can also consider that the $(N\times K)$-dimensional matrix $\mathbf{X}$ as the input to a CNN without word embedding layer. On the other hand, the second approach is to train $\mathbf{W}_{e}$ along with the other parameters of CNN. The initial values of $\mathbf{W}_{e}$ can be either randomly generated or set to be the pre-trained universal vectors such as those from Word2vec. In this paper, we adopt the  second approach and train $\mathbf{W}_{e}$ so that its values are specific to the patent data set.\par

\paragraph{Convolutional layer}
The purpose of the convolutional layer is to mine the sentence to grasp a truly abstract representations comprising rich semantic information. Since a word embedding vector is an integral entity that is meaningless when divided apart, one-dimensional convolutional is commonly used for NLP instead of the two-dimensional convolutional for image processing. Specifically, the width of the filter is fixed as $K$, which is the same as the length of the word embedding vector. The filters only slide through the $(N\times K)$-dimensional matrix $\mathbf{X}$ along its longitudinal direction instead of along both longitudinal and transverse directions as shown in Fig. \ref{cnn}. The length $\lambda$ of the filters can be set to different values, which has physical implications as $\lambda$ terms are convoluted together and their semantics are extracted as a whole to keep better context information. Generally speaking, larger values of $\lambda$ correspond to longer filter length, and therefore can preserve more context information by convoluting larger number of terms at the same time. On the other hand, the value of $\lambda$ should not be too large as this may compromise the word-level semantic information of the word embedding vectors, and also lead to large computation complexity.\par

Let $\mathbf{W}_{c}\in\mathbb{R}^{\lambda\times K}$ denote the $(\lambda\times K)$-dimensional weight matrix between the word embedding layer and the convolutional layer, and $b$ denote the bias value. Therefore, for the block of matrix $\mathbf{X}$ between the $\alpha$-th row and $(\alpha+\lambda-1)$-th row, i.e., $\mathbf{X}_{\alpha:\alpha+\lambda-1}$, the extracted feature value $c_{\alpha}$ can be obtained as follows:
\begin{equation}
\label{eq1}
c_{\alpha}=\sigma(\mathbf{W}_{c}\otimes\mathbf{X}_{\alpha:\alpha+\lambda-1}+b),
\end{equation}
\noindent where $\otimes$ means the convolution of two matrixes and $\sigma$ is the activation functions such as Sigmoid, Tanh, ReLU, etc.. As the filter slides through the matrix $\mathbf{X}$ with a step size of $1$, an $(n-\lambda+1)$-dimensional feature vector can be generated as $\mathbf{c}=[c_{1},c_{2},\cdots,c_{\alpha},\cdots,c_{N-\lambda+1}]$.\par

Generally speaking, there can be multiple filters in a CNN and the size of the filters can be different. Consider there are $\theta$ filter sizes denoted by $\{\lambda_{1},\cdots,\lambda_{\theta}\}$, and $\mu$ filters per filter size. Therefore, we can obtain $\mu\times\theta$ feature vectors $\{\mathbf{c}_{i,j}|i\in\theta,j\in\mu\}$. Note that the feature vectors obtained by filters with different sizes have different dimensions, and the dimension for a feature vector $\mathbf{c}_{i,j}$ is $N-\lambda_{i}+1$.\par

\paragraph{Pooling layer}
The purpose of the pooling layer is twofold. Firstly, the feature resolution is reduced to avoid overfitting. Secondly, it solves the problem that the convolutional layer feature vector dimensions are different for different patent documents. This is because that the dimensions of the feature vectors generated by the convolutional layer as discussed above depend on the number of terms $N$ in a patent document (i.e., $N-\lambda+1$ for a filter with length $\lambda$), while the value of $N$ varies from patent to patent. As pooling selects a fixed number of elements from a convolutional layer feature vector to construct a new feature vector, this guarantees that the dimension of the pooling layer feature vector is independent of the value of $N$. \par

There are commonly two pooling strategies, i.e., maximum (MAX) pooling and average pooling. As MAX pooling demonstrates better performance for NLP \cite{zhang2017sensitivity}, we adopt MAX pooling in this paper. Consider that the $\omega$ largest elements are selected from a convolutional layer feature vector $\mathbf{c}=[c_{1},c_{2},\cdots,c_{\alpha},\cdots,c_{N-\lambda+1}]$ to construct a new vector, i.e.,
\begin{equation}
\label{eq2}
\mathbf{\hat{c}}=[c_{1},c_{2},\cdots,c_{i},\cdots,c_{\omega}], \ i\in\{1,\cdots,\omega\}
\end{equation}
\noindent where $c_{i}$ is the $i$-th largest element in $\mathbf{c}$. As there are $\mu\times\theta$ feature vectors $\{\mathbf{c}_{i,j}|i\in\theta,j\in\mu\}$, we denote the set of corresponding new vectors by $\{\mathbf{\hat{c}}_{i,j}|i\in\theta,j\in\mu\}$. The pooling layer feature vector is the concatenation of all new vectors, i.e.,
\begin{equation}
\label{eq3}
\mathbf{\hat{C}}=[\mathbf{\hat{c}}_{1,1},\cdots,\mathbf{\hat{c}}_{1,\mu},\cdots,\mathbf{\hat{c}}_{\theta,1},\cdots,\mathbf{\hat{c}}_{\theta,\mu}],
\end{equation}
\noindent which has a dimension of $\omega\times\theta\times\mu$. The above pooling layer feature vector in \eqref{eq3} is the feature vector of a patent used to construct the FVSM.\par

\paragraph{Fully-connected layer}
The pooling layer feature vector is used as input to the fully-connected layer, whose output is the classification result of the corresponding patent. A linear activation function is used for the fully-connected layer. Specifically, let $m$ denote the total number of classes for the patents, and $\mathbf{W}_{l}\in\mathbb{R}^{m\times(\omega\times\theta\times\mu)}$ denote the weight matrix for the fully-connected layer, and $\mathbf{b}_{l}\in\mathbb{R}^{m\times 1}$ denote the bias vector. Therefore, we can obtain the vector $\mathbf{y}\in\mathbb{R}^{m\times 1}$ according to
\begin{equation}
\label{eq4}
\mathbf{y}=\mathbf{W}_{l}\times\mathbf{\hat{C}}^{\mathrm{T}}+\mathbf{b}_{l},
\end{equation}
\noindent and use Softmax to convert $\mathbf{y}$ to the probabilities that a patent belongs to different classes, which is the output of fully-connected layer and CNN, i.e.,
\begin{equation}
\label{eq5}
\mathbf{\hat{y}}=\mathrm{Softmax}\{\mathbf{y}\},
\end{equation}

\subsection{FVSM Application}
\subsubsection{Patent similarity}
The first application of our proposed FVSM for patent analysis is to assess patent similarity, which is also a typical application for VSM. For this purpose, we first design a number of patent triads with each triad consisting of three patents as a group, namely $\mathbf{P}$, $\mathbf{P}^{+}$ and $\mathbf{P}^{-}$. $\mathbf{P}$ denotes a base patent, while the other two patents $\mathbf{P}^{+}$ and $\mathbf{P}^{-}$ are provided for comparison with the base patent $\mathbf{P}$. The difference between $\mathbf{P}^{+}$ and $\mathbf{P}^{-}$ is that the former is more similar to $\mathbf{P}$ than the latter.\par

To quantify the patent similarity between any two patents based on FVSM, we can use three measures, namely, Euclidean distance $d(\mathbf{\hat{C}}(\mathbf{P}), \mathbf{\hat{C}}(\mathbf{P}^{+}))$, Cosine of the angular separation $\cos(\theta(\mathbf{\hat{C}}(\mathbf{P}), \mathbf{\hat{C}}(\mathbf{P}^{+})))$ (i,e, Cosine similarity), and Jaccard index $J(\mathbf{\hat{C}}(\mathbf{P}), \mathbf{\hat{C}}(\mathbf{P}^{+}))$ (i,e, Jaccard similarity) between the feature vectors of the two patents, where the feature vector $\mathbf{\hat{C}}(\mathbf{P})$ of patent $\mathbf{P}$ is derived according to \eqref{eq3}. Let $p$ be a general notation for any of the three measures $d$, $\cos(\theta)$ and $J$. Therefore, for each patent triad, we get two measure values, one for the similarity between $\mathbf{P}$ and $\mathbf{P}^{+}$, i.e., $p(\mathbf{\hat{C}}(\mathbf{P}), \mathbf{\hat{C}}(\mathbf{P}^{+}))$ and the other one for the similarity between $\mathbf{P}$ and $\mathbf{P}^{-}$, i.e., $p(\mathbf{\hat{C}}(\mathbf{P}), \mathbf{\hat{C}}(\mathbf{P}^{-}))$. Therefore the situation where  $p(\mathbf{\hat{C}}(\mathbf{P}), \mathbf{\hat{C}}(\mathbf{P}^{+}))<p(\mathbf{\hat{C}}(\mathbf{P}), \mathbf{\hat{C}}(\mathbf{P}^{-}))$ means that the result of patent similarity discrimination given by FVSM is consistent with that given by manual labeling. Finally for all the patent triads, we use the ratio between the number of consensus rating with the total number of ratings, i.e.,  the accuracy rate of patent similarity discrimination, as an indicator for the accuracy of FVSM.  \par

\subsubsection{Patent clustering}
Patent clustering is another application of the proposed FVSM for patent analysis. The clustering can be performed by K-means method, which is an unsupervised learning algorithm having advantages of succinct operation and efficient calculation to obtain reasonable clustering results. The accuracy of patent clustering using the K-means method depends on the choice of cluster number. Usually, there are two common methods for finding the appropriate number of clusters within a dataset, namely elbow method and contour coefficient method. In practical applications, we can select the number of clusters produced by either of them as the reference. An excellent FVSM or VSM generated by a patent analysis model will lead to a situation, where feature vectors symbolizing a group of similar patents tend to be clustered together. The group of patents are referred to as a patent cluster. We could forecast a vacant area of a given technology field by clustering patent documents and identify clusters with smaller number of patent documents as potential future technologies. \par

\subsubsection{Patent map}
Although FVSM and VSM can be used for patent analysis as discussed above, it is difficult to visualize the constructed spaces due to their high-dimensionality. Therefore, we use dimensionality reduction methods to generate a patent map and visualize the clustering result by K-means. Generally speaking, there are two broad categories of dimensionality reduction methods, one is the linear methods such as Principal Component Analysis (PCA), while the other is the non-linear methods such as T-distributed Stochastic Neighbor Embedding (T-SNE). Compared with linear methods, non-linear methods generally have the advantages that the resulting images spread the distribution of each category, making the boundaries of each category clearer. 

In this paper, we reduce all the patent feature vectors to two dimensions and generate a two-dimensionality patent map. We use different colors to represent different patent clusters generated by K-means, so that we can visualize the cluster distributions in the patent map. For example, a technology can be considered as relatively independent if the corresponding patent clusters are far away from the other patent clusters, and vice versa. Moreover, two technologies can be considered as closely related if there have overlapping areas between the two corresponding patent clusters. \par

\section{Case Study: Internet-of-Things}
In this section, the patents related to Internet of things (IoT) technology are chosen to be analyzed as a case study in order to demonstrate the effectiveness and reliability of the patent analytics based on FVSM. IoT comprises smart devices with sensors, actuators, and communication modules that are connected to the Internet. It is a new revolution to the traditional Internet, which not only interconnects people but also ``things", i.e., smart devices. IoT is expected to have significant applications in industry, agriculture, medical, transportation, etc., and to have huge impact on the world's economy and quality of life. IoT includes a group of enabling technologies that can be divided into identification, sensing, communications, computing, service, and semantics \cite{al2015internet}. Therefore, identifying innovation patents of IoT technology and analyzing their development trends is with great importance for promoting the new IoT development. \par

Our investigation on IoT patents are mainly carried out through three steps, i.e., data collection and pre-processing, FVSM construction and FVSM application. In the last step, a quantitative evaluation is established for each application through specific score results. The details of each step are given as follows.\par

\subsection{Data Collection and Preprocessing}
Firstly, the IoT related patents have to be collected as experimental data sets. We use the patent data from year 2016 to 2018 in United States Patent \& Trademark Office (USPTO). USPTO maintains a freely available database of patent data, where each patent is classified in one or multiple International Patent Classification (IPC) classes \cite{USPTO2018}. Only the patents under the IPC class of H04 are included for our study because this class covers most of IoT related technologies. Before filtering, all the patents in H04 have to be processed to remove the stop word and stem the words by Porter stemming in order to remove noise in the corpus \cite{ali2012porter}. Then, based on the stemmed IoT-related keywords and phrases, some of which are presented in Table. \ref{keywords} for the sake of illustration, all the patents in H04 are filtered to generate the patent data set for experiments. Specifically, if a patent has a keyword or phrase as defined, it is included in the data set. In this way, 8942 patents are selected and their patent texts constitute a structured patent data set.\par
\newcommand{\tabincell}[2]{\begin{tabular}{@{}#1@{}}#2\end{tabular}} 
\begin{table}
	\centering
	\caption{Samples of processed keywords and phrases for filtering IoT patent data set}
	\setlength{\tabcolsep}{3pt}
	\begin{tabular}{|c|c|c|} 
		\hline
		'internet','thing'&'iot'\\
		\hline
		'rfid','tag'&'smart','applianc'\\
		\hline
		'smart','devic'&'rfid'\\
		\hline
		'm2m'&'machin','machin'\\
		\hline
		'nb','iot'&'bluetooth'\\
		\hline
		'zigbe'&'ble'\\
		\hline
		'lowpan'&'bluetooth','low','energi'\\
		\hline
		'pda'&'wearabl','devic'\\
		\hline
		'person','digit','assist'&'smart','home'\\
		\hline
		'cloud','comput'&'cloud','platform'\\	
		\hline
		'sensor'&'smart','grid'\\
		\hline
		'embed','system'&'smart','citi'\\
		\hline
		'autom'&'wsn'\\
		\hline
		'healthcar'&'health','care'\\
		\hline
		'3g'&'4g'\\
		\hline
		'near','field'&'nfc'\\
		\hline
		'gsm'&'lte'\\
		\hline
		'cdma'&'wcdma'\\
		\hline
		'802', '11'&'802', '15', '4'\\
		\hline
		'wimax'&'wifi'\\
		\hline
		'ipv6'&'ipv4'\\
		\hline
		'6lowpan'&'fpga'\\
		\hline
		'android'&'actuat'\\
		\hline
		'cloud','base','server'&'cloud', 'server'\\
		\hline
		'epc'&'big','data'\\
		\hline
		'lora'&'authentic'\\
		\hline 
	\end{tabular}
	\label{keywords}
\end{table}

In order to extract feature vectors of the above patents using CNN, we need to first label the patents according to their classification results, based on which the CNN can be trained. Since the number of patents in the data set is too large, manually labeling the patents is not practical. Instead, we use LDA method to solve the labeling problem. LDA is a generative probabilistic model for collections of discrete data such as text corpora, in which each item of a collection is modeled as a finite mixture over an underlying set of topics. Each topic can be modeled as an infinite mixture over an underlying set of topic probabilities. By using LDA, we can conveniently calculate what topics are contained in the corpus and what keywords are contained in a topic in an unsupervised manner while taking the semantic information in the corpus into account. We compose all the patent texts in the data set into a corpus, and use the LDA model to extract a given number of topics, e.g., 8, each of which can be regarded as a category. Based on the trained LDA model, we can calculate the patent-topic probability matrix, where each row vector indicates probability distribution of topics to a patent. The topic with the highest probability value in each patent can be regarded as its category, thus completing the labeling of the patent data set. In a nutshell, the IoT-related patents are classified according to their topics, and the topic of a patent is determined to be the one with the largest probability according to the LDA.\par

	\newtheorem{remark}{Remark}
\begin{remark}[Discussion on patent labeling methods]	
	In this paper, we apply LDA to derive the topics of the patents, which are used as the classification results. A simpler and more straightforward method is to use the IPC class number for classification. IPC provides a hierarchical system of language independent symbols for the classification of patents and utility models according to the different areas of technology to which they pertain \cite{jun2011ipc}. Characters from an IPC code represent the field in which the patent belongs to. Every IPC code includes some hierarchies referred to as sections, classes, subclasses and so on from high-level to low-level. For example, the patents in our data set come from the IPC class H04, where "H" is the section representing the Electricity and "04" is the class representing the Electric Communication Technique. We adopt the subclasses of IPC code as the patent labels. We found that the patents selected for the IoT data set contain 11 subclasses in the IPC class H04. In other words, these patents belong to 11 categories. There are some advantages of using the IPC code as a category label such as simplicity and convenience. However, the relation between the category label and the text content of a patent is not obvious and convincing. If the training of CNN is based on the labels by IPC class number, the reliability of the FVSM obtained is not satisfactory. For example, in our experiments, it is found that the CNN classification accuracy rate fluctuates between $40\%$ and $50\%$ after training, which means that such labeling method is not a good option. Note that theoretically, the classification results obtained by any patent classification methods can be used to train the CNN.
\end{remark}

\subsection{FVSM Construction}
\begin{figure*}
	\centering
	\includegraphics[width=1.05\textwidth]{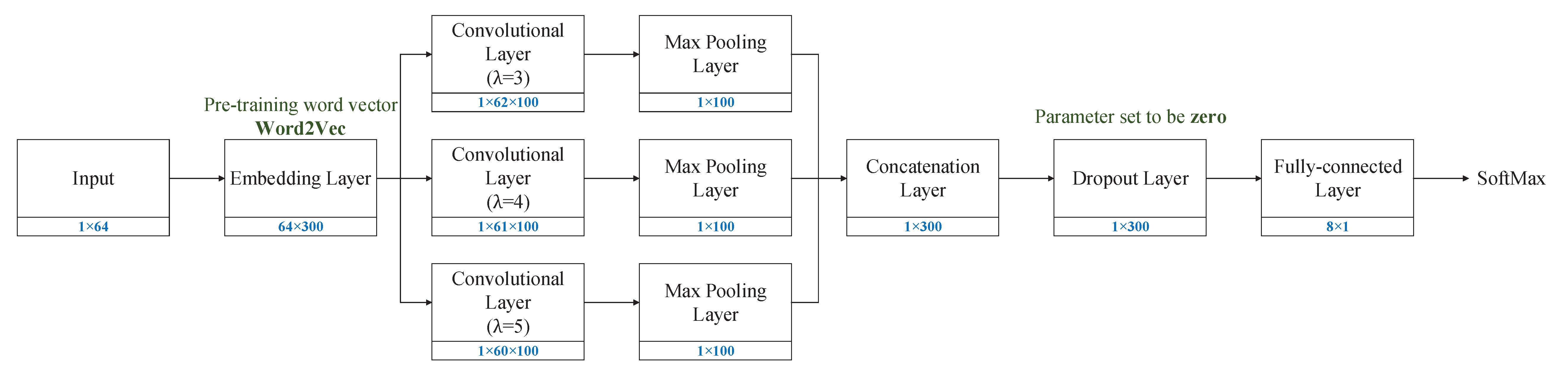}
	\caption{CNN used for FVSM construction in our experiments}
	\label{casestudyCNN}
\end{figure*}

Fig. \ref{casestudyCNN} illustrates the flow chart for training CNN and extract feature vectors to construct the FVSM used in the following applications. Here detailed information on the CNN hyper-parameters and training process are presented, including the training method of the word embedding layer, the number and size of the convolution kernel, the pooling strategy, etc.\par

The weights of word embedding layer are initialized with the pre-trained word vectors given by Word2Vec \cite{goldberg2014word2vec}. Then, the weights of this layer is fine-tuned during the training process. In the convolutional layer, we set up $\theta=3$ types of convolution kernels with the sizes of $3$, $4$, and $5$, respectively. The number of convolution kernels with the same size is set to $\mu=100$. Thus, after convolutional layer, a total of $\theta\times\mu=300$ convolutional layer feature vectors can be captured. In the pooling layer, the maximum pooling strategy is used to extract the largest value from each convolutional layer feature vector as output. After splicing each value which come from the output of the pooling layer, we obtain the $300$-dimensional feature vector as the input into the classifier.\par

For the classifier, the dropout parameters need to be set to zero, which means that no data is discarded. This is because that our objective is to preserve the $300$-dimensional feature vector as the training result without losing the information contained in any dimension. Although there exists the possibility of overfitting, all the information in the feature vectors can be reserved, which is essential to well exploit the FVSM and keep its integrity. Finally, the feature vectors pass through a fully connected layer and the probability of a given number of categories are generated as output.\par

In order to calculate the accuracy of the classification results, we adopt the $l$-fold cross-validation with $l=10$. Cross-validation is a statistical method used to estimate the skill of machine learning models. Specifically, some patent samples in the data set are first selected as the test set. Then, the rest of the samples are randomly partitioned into $l$ equal size subsamples. Of the $l$ subsamples, a single subsample is retained as the validation data for testing the model, and the remaining $l-1$ subsamples are used as training data. In the training process, random gradient descent algorithm with update rules of AdaDelta and a mini-batch size of $50$ is used to train the neural network. The cross-validation process is then repeated $l-1$ times, with a different subsample chosen as the validation data in each time. Finally, all the trained CNNs are emulated by using the samples in the test set.\par

The results of classification accuracy of the CNN are shown in Table. \ref{CNN result}. The average accuracy of the validation set is up to $88.4\%$ while the average accuracy of the test set is up to $88.0\%$. The weights of the CNN with the highest accuracy of the test set are chosen for our further application, i.e., from the $2$nd fold. In other words, the corresponding output of the pooling layer is taken as the feature vector of the patent.\par
\begin{table*}
	\normalsize
	\centering
	\caption{Scores of cross-validation}
	\setlength{\tabcolsep}{3pt}
	\begin{tabular}{|c|c|c|c|c|c|c|c|c|c|c|c|} 
		\hline 
		Fold Index&1&2&3&4&5&6&7&8&9&10&Average\\
		\hline  
		Validation Scores&0.91&0.89&0.876&0.864&0.875&0.893&0.895&0.884&0.875&0.882&0.884\\
		\hline 
		Test Scores&0.859&0.917&0.898&0.904&0.902&0.877&0.886&0.881&0.851&0.825&0.880\\
		\hline 
	\end{tabular}
	\label{CNN result}
\end{table*}

\subsection{FVSM Applications}
\subsubsection{Patent Similarity}
More than $200$ patents are manually selected from the patent data set to construct the set of patent triads, i.e., $\{\mathbf{P}, \mathbf{P}^{+}, \mathbf{P}^{-}\}$. As discussed in Section III-B-1), we derive the patent similarity measure values between patent pair $(\mathbf{P}, \mathbf{P}^{+})$ and patent pair $(\mathbf{P}, \mathbf{P}^{-})$ based on FVSM. By comparing the two similarity measure values, we can determine which patent, i.e., $\mathbf{P}^{+}$ or $\mathbf{P}^{-}$, is more similar to the base patent $\mathbf{P}$ for each patent triad. The discrimination results are compared with the manual labeling results for all the patent triads to derive the accuracy rate.\par 

According to the different levels of ``labeling difficulty", we can divide the labeled patent triads into two sets, i.e., $\mathbb{S}_{1}$ and $\mathbb{S}_{2}$. The ``labeling difficulty" is lower for the former while higher for the latter. Specifically in $\mathbb{S}_{1}$, there is a clear similarity between the topics and keywords of $\mathbf{P}$ and $\mathbf{P}^{+}$, while the topics and keywords of $\mathbf{P}^{-}$ are obviously different from those of $\mathbf{P}$ and $\mathbf{P}^{+}$. In $\mathbb{S}_{2}$, there is not much difference between the topics and keywords of $\mathbf{P}$ , $\mathbf{P}^{+}$ and $\mathbf{P}^{-}$. An example that explains the difference between $\mathbb{S}_{1}$ and $\mathbb{S}_{2}$ is given in Table. \ref{traid example}.\par
\begin{table}
	\normalsize
	\centering
	\caption{Examples of keywords in patent triads}
	\setlength{\tabcolsep}{3pt}
	\begin{tabular}{|c|c|c|c|} 
		\hline 
		&$\mathbf{P}$&$\mathbf{P}^{+}$&$\mathbf{P}^{-}$\\
		\hline  
		$\mathbb{S}_{1}$&\tabincell{c}{wearable device,\\ sensor}&\tabincell{c}{wearable device,\\ RFID}&5G\\
		\hline 
		$\mathbb{S}_{2}$&\tabincell{c}{wearable device,\\ sensor}&\tabincell{c}{wearable device,\\ RFID}&\tabincell{c}{wearable device,\\ authenticate}\\
		\hline 
	\end{tabular}
	\label{traid example}
\end{table}

Based on the aforementioned labeling strategy, we produce two sets of patent triads, i.e., $\mathbb{S}_{1}$ with $156$ patent triads, and $\mathbb{S}_{2}$ with $61$ patent triads. Then, the accuracy rates of patent similarity discrimination based on the proposed FVSM in terms of Euclidean distance as discussed in Section III-B on the two data sets are derived as shown in Table. \ref{traid result}. For comparison purposes, we also derived the accuracy rates of patent similarity discrimination based on  TF-IDF-based VSM.\par
\begin{table}	
    \normalsize
    \centering
    \caption{Accuracy rate in terms of Euclidean distance on benchmark patent data sets}
    \setlength{\tabcolsep}{3pt}
    \begin{tabular}{|c|c|c|} 
	    \hline 
	    \diagbox[width=3cm]{Data Set}{Accuracy Rate}&Source of VSM&\tabincell{c}{Euclidean \\Distance}\\
	    \hline  
	    \multirow{2}*{$\mathbb{S}_{1}$}&CNN&{\bfseries 91.0\%}\\
	    \cline{2-3}
	    &TF-IDF&82.1\%\\
	    \hline 
	    \multirow{2}*{$\mathbb{S}_{2}$}&CNN&{\bfseries 67.2\%}\\
	    \cline{2-3}
	    &TF-IDF&63.9\%\\
	    \hline
	    \multirow{2}*{$\mathbb{S}_{1}\cup\mathbb{S}_{2}$}&CNN&{\bfseries 84.3\%}\\
	    \cline{2-3}
	    &TF-IDF&77.0\%\\
	    \hline 
    \end{tabular}
	\label{traid result}
\end{table}

As expected, the accuracy rates of the proposed FVSM are consistently higher than those of TF-IDF-based VSM for both $\mathbb{S}_{1}$ and $\mathbb{S}_{2}$. It can be seen that the FVSM method can reliably determine the similarity between patents. In addition, the accuracy rates of FVSM for $\mathbb{S}_{1}$ is higher than that for $\mathbb{S}_{2}$, i.e., $91.0\%$, and $67.2\%$. This is because the former has a smaller ``labeling difficulty" when designing the data set.

\subsubsection{Patent clustering}
We first find the appropriate number of clusters $\kappa$ using the elbow method. With increasing $\kappa$, the descending rate of SSE (Sum of Squares for Error) becomes slower and slower in the range of $5$ to $50$ clusters. It can be found that the curve has the highest curvature when $\kappa=18$, i.e., the optimal value of $\kappa$ is 18.\par
Thus, using the K-means method, we divided all patents into 18 clusters. Next, in each cluster, LDA is used to extract the representative keywords contained in all the patents to infer a definition for the technologies in that cluster. We adopt a method similar to that in \cite{Kim2015}, where the top five keywords in the top two topics are used as the representative keywords for a patent cluster. The representative keywords contained in patent text of all the clusters with $\kappa=18$ are given in Table. \ref{cluster} (due to the space limitation, only the first 8 keywords of each cluster are shown here). Several clusters are chosen for illustration. For example, the second cluster contains keywords such as "rfid", "tag", "wireless", "nfc", etc., which means that the patents in this cluster are mostly related to RFID tag. The tenth cluster contains keywords such as "antenna", "power", "signal", "phase", etc., which means that the patents in this cluster are mostly related to radio frequency.\par

Therefore, it can be seen that the FVSM-based clustering analysis can well classify the IoT-related patents into different technology clusters, which makes it easier for researchers to discover the distribution of patents for various sub-technologies. In order to better visualize the patent distribution for IoT technology, we generate a two-dimensional patent map which is discussed as below.

\begin{table*}	
	\small
	\centering
	\caption{Keywords of IoT-related patent clusters}
	\setlength{\tabcolsep}{3pt}
	\begin{tabular}{|c|c|c|c|c|c|c|c|c|c|} 
\hline 
\diagbox[width=2.2cm]{\textbf{Cluster}}{\textbf{Keywords}}&Word1&Word2&Word3&Word4&Word5&Word6&Word7&Word8&\tabincell{c}{Technology \\Field}\\
\hline 
1&optic&voltag&circuit&power&clock&noise&wavelength&current&\tabincell{c}{IoT \\hardware}\\
\hline 
2&rfid&tag&wireless&wifi&nfc&node&mobile&network&\tabincell{c}{RFID \\tag}\\
\hline 
3&light&image&pixel&gravity&speed&filter&wavelength&sound&sensor\\
\hline 
4&node&packet&switch&traffic&port&address&ip&path&\tabincell{c}{network \\optimization}\\
\hline 
5&terminal&node&remote&connect&module&unit&network&access&\tabincell{c}{access \\control}\\
\hline 
6&inform&unit&data&terminal&compress&value&decode&client&IoT terminal\\
\hline 
7&network&application&resource&content&address&service&subscibe&access&web service\\
\hline 
8&security&call&authenticate&encrypt&subscibe&key&token&route&\tabincell{c}{network \\security}\\
\hline 
9&locate&compute&geography&stream&signal&multimedia&mobile&contain&\tabincell{c}{stream \\media}\\
\hline 
10&antenna&power&signal&switch&channel&interfer&transmiss&phase&\tabincell{c}{radio \\frequency}\\
\hline 
11&audio&stream&video&sensor&sound&broadcast&track&signal&audio\\
\hline 
12&network&mobile&wireless&operation&interface&access&control&electronic&\tabincell{c}{wireless \\nerwork \\access}\\
\hline 
13&resource&band&carrier&network&base&station&uplink&frequence&\tabincell{c}{spectrum \\optimization}\\
\hline 
14&image&pixel&display&color&section&screen&light&region&\tabincell{c}{image \\processing}\\
\hline 
15&cell&downlink&radio&quality&csi&interfer&neighbor&ack&\tabincell{c}{mobile \\network}\\
\hline 
16&ofdm&sample&encode&filter&phase&digit&interfer&estim&\tabincell{c}{signal \\processing}\\
\hline 17&media&stream&video&server&request&television&electronic&virtual&\tabincell{c}{video \\transmission}\\
\hline 
18&display&touch&light&screen&sensor&detect&control&surface&\tabincell{c}{IoT \\peripheral}\\
\hline 
    \end{tabular}
    \label{cluster}
\end{table*}

\subsubsection{Patent map}
\begin{figure}
	\centering
	\subfigure[by PCA]{
		\label{figa} 
		\includegraphics[width=8cm]{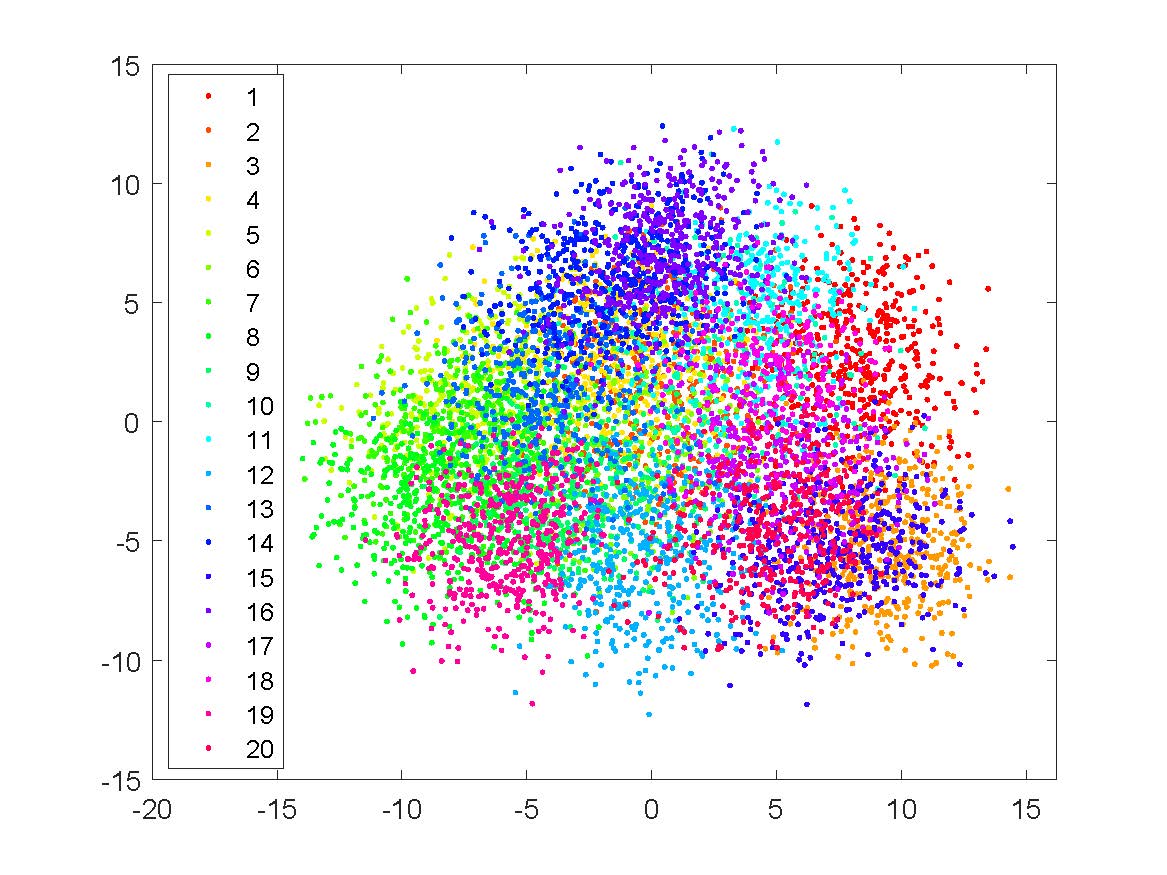}}
	\hspace{1in}
	\subfigure[by T-SNE]{
		\label{figb} 
		\includegraphics[width=8cm]{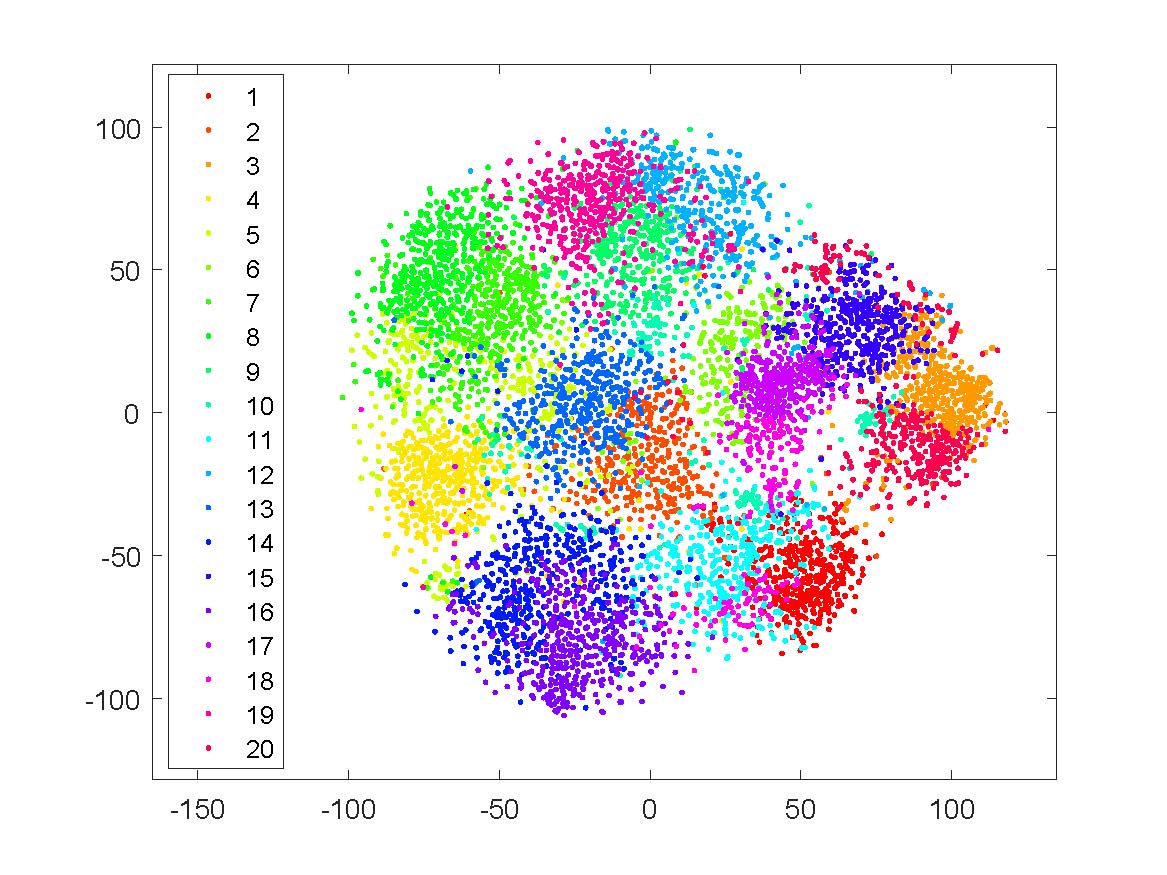}}
	\caption{FVSM Results by Dimensionality Reduction}
	\label{patentmap} 
\end{figure}

The K-means clustering results of IoT-related patents can be visualized by generating the patent maps. We first apply a linear dimensionality reduction algorithm, i.e., PCA. We reduce all patent feature vectors into two-dimensional using PCA, and use different colors to represent different clusters to which the patents belong, resulting in a patent map as shown in Fig. \ref{figa}. Although the patent map obtained by the PCA algorithm can show that all the feature vectors are roughly gathered in several clusters, the distribution of various patents in this map is rather messy and even overlap in some places. Therefore, in order to obtain a patent map with better visualization effect, a non-linear dimensionality reduction algorithm, i.e., T-SNE, is used. As shown in Fig. \ref{figb}, it can be found that there is a clearer border between different clusters in the patent map obtained by using the T-SNE algorithm. This also proves from another aspect that the clustering results obtained by the aforementioned K-means method are convincing.\par

\begin{figure*}
	\centering
	\includegraphics[width=0.9\textwidth]{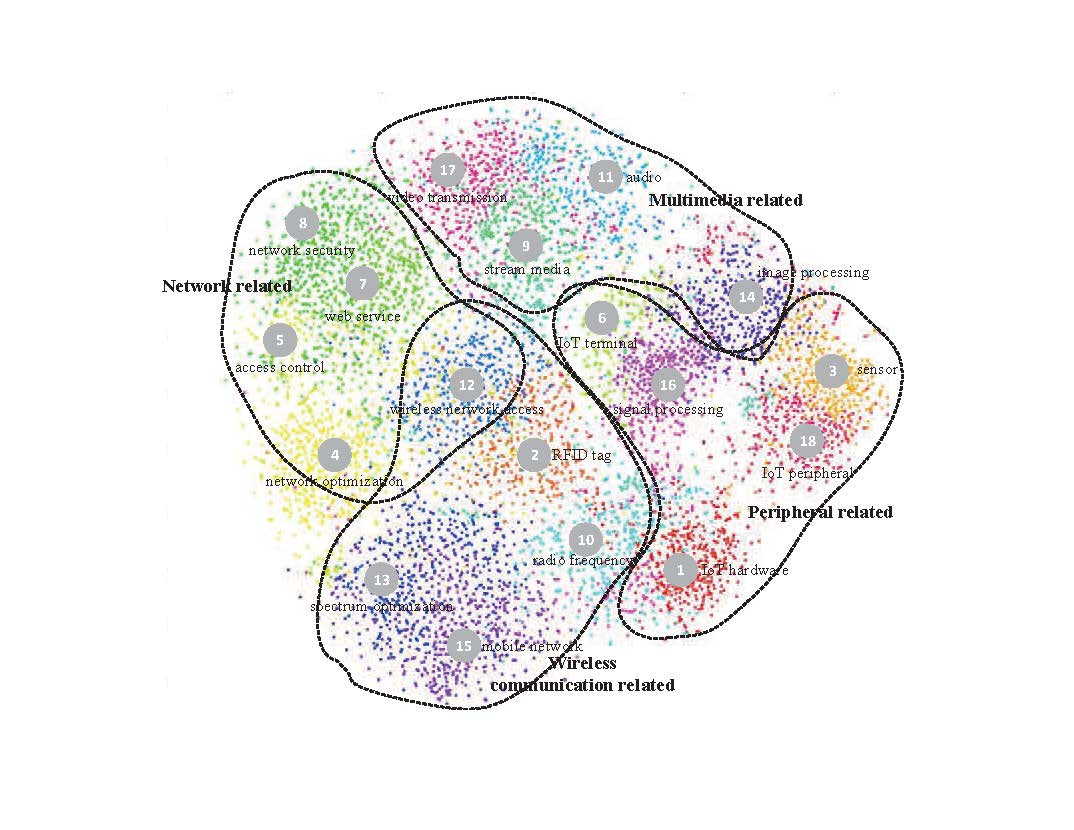}
	\caption{IoT patent map with technology fields}
	\label{tsnefigure}
	\vspace{-1em}
\end{figure*}

Based on Fig. \ref{figb} and Table. \ref{cluster}, we plot the IoT patent map in Fig. \ref{tsnefigure}. It can be found that some closely-related technology fields tend to cluster together in the patent map, such as \textbf{audio}, \textbf{stream media} and \textbf{image processing}, which may form a greater \textbf{supercluster}, such as \textbf{multimedia related}. It is interesting to notice that these superclusters are in accordance with the IoT architeture \cite{al2015internet}, e.g., peripheral related supercluster corresponds to the perception layer, wireless communications and network related superclusters correspond to network layer, etc.. Moreover, some technology fields lie in the overlapping area and belong to different superclusters, such as the \textbf{wireless network access} cluster. This indicates that patents in those fields are related to more than one superclusters. By using patent maps, we can quantitatively represent specific patents and accurately locate them on the map, so as to analyze the situation of patents more effectively.

\section{Conclusion}
In this paper, a feature vector space model based on deep learning is proposed to extract features from patent texts. We have not only proposed a patent analysis method, but also applied it to analyse the patents in the IoT technology field. Based on a massive number of patents in IPC H04 of USPTO patent database, we have established a huge patent data set through text de-noising such as Porter stemming. Based on this data set, the validation of the proposed patent analytics is fully proved by the study on the IoT related patents. In terms of comparing patent similarities, it is shown that an accuracy of up to about $90\%$ can be achieved. At the same time, using T-SNE as a dimensionality reduction method, we have clearly classified and visualized IoT related patents. In the next steps, we will further improve the neural network to achieve higher accuracy and generality.\par

\section*{Acknowledgment}

This work was supported by the China Natural Science Funding under the grant 61671089, and BUPT Excellent Ph.D. Students Foundation under grant XTCX201828
.

\bibliography{author}
\bibliographystyle{IEEEtran}


\end{document}